\documentclass[11pt]{article}
\usepackage{ranlp97}

\newcommand{\amalia}{\mbox{$\cal{A}${\sc malia}}}
\newcommand{\ale}{\mbox{\sc ale}}
\newcommand{\tag}[1]{\fbox{\scriptsize #1}}
\newenvironment{tfs}[1]{\left[\!\!\!\!\!\begin{array}{ll}{\mbox{\bf
#1}}\\}{\end{array}\!\!\!\!\!\right]}
\title{\amalia\ -- A Unified Platform for Parsing and
  Generation\thanks{\ \ In R.\ Mitkov, N.\ Nicolov and N.\ Nikolov,
  eds., {\em Proceedings of
``Recent Advances in Natural Language Processing'' (RANLP'97)}, pp. 135-142,
Tzigov Chark, Bulgaria,
11-13 September 1997}}
\author{Shuly Wintner\thanks{\ \ Supported by the Minerva
Stipendien Komitee.} \\
Seminar f\"ur Sprachwissenschaft\\
Universit\"at T\"ubingen\\
Kl.\ Wilhelmstr.\ 113\\
72074 T\"ubingen, Germany\\
{\tt shuly@sfs.nphil.uni-tuebingen.de}
\And Evgeniy Gabrilovich 
\and Nissim Francez\thanks{\ \ Supported by a grant from the 
Israeli Ministry of Science: ``Programming Languages Induced
Computational Linguistics'' and the Fund for the Promotion of 
Research in the Technion. We thank an anonymous reviewer for 
useful comments.}\\
Laboratory for Computational Linguistics\\
        Computer Science Department\\
        Technion, Israel Institute of Technology\\
        32000 Haifa, Israel\\
        {\tt \{gabr,francez\}@cs.technion.ac.il}}

\begin{document}
\maketitle

\begin{abstract}
Contemporary linguistic theories (in particular, HPSG) are declarative
in nature: they specify constraints on permissible structures, not how
such structures are to be computed. Grammars designed under such
theories are, therefore, suitable for both parsing and generation.
However, practical implementations of such theories don't usually
support bidirectional processing of grammars. We present a grammar
development system that includes a compiler of grammars (for parsing
and generation) to abstract machine instructions, and an interpreter
for the abstract machine language. The generation compiler inverts
input grammars (designed for parsing) to a form more suitable for
generation. The compiled grammars are then executed by the interpreter
using one control strategy, regardless of whether the grammar is the
original or the inverted version.  We thus obtain a unified, efficient
platform for developing reversible grammars.
\end{abstract}

\section{Introduction}
The popularity of contemporary linguistic formalisms such as Lexical
Functional Grammar \cite{lfg}, Categorial Grammar \cite{cg} or
Head-Driven Phrase-Structure Grammar (HPSG)~\cite{hpsg2}, and
especially their mathematical and formal maturity, have led to the
development of various frameworks, applying different methods, for
their implementation.

This paper focuses on a computational framework in which HPSG grammars
can be developed. A wide spectrum of implementation techniques for
HPSG exist: one extreme is direct interpretation of grammars. For
parsing, this involves a program that accepts as input a grammar and a
string and parses the string according to the grammar. For generation,
the input is a semantic formula from which a phrase is generated. The
earliest HPSG parsers (e.g.,~\cite{prupol85,franz}) were designed in
this way. A slightly more elaborate technique is the use of some
high-level, unification-based logic programming language (e.g., Prolog
or LIFE~\cite{life-meaning}) for specifying the grammar. Further along
this line lies compilation of grammars directly {\em into\/} Prolog,
using Prolog's internal mechanisms for performing unification. This is
the implementation technique of, e.g., Profit~\cite{profit}. Systems
such as \ale~\cite{ale,carp95} also compile grammars into
Prolog. However, \ale\ compiles grammar descriptions directly into
Prolog code, rather than into (a Prolog representation of) feature
structures. At run time, \ale\ executes the code that was compiled for
the rules. Parts of the unifications (resulting from type-unification)
are performed at compile-time to increase the efficiency of the
generated code.

In this paper we advocate a further step along the same spectrum. We
propose \amalia, an abstract machine specifically designed for
executing \ale\ grammars (without relational extensions).  \amalia\
includes a compiler of input grammars into the abstract machine
language and an interpreter for the abstract instructions. This
implementation technique was proved useful for many programming
languages, most notably Prolog itself\footnote{Recently, such
techniques were used for implementing the new programming language
Java.}, and as we show below, it improves the efficiency of parsing
with \ale\ grammars considerably. We emphasize in this paper the more
practical aspects of the system, focusing on the integration of
parsing and generation, as its theoretical infra-structure has been
presented elsewhere~\cite{shuly:nlulp-95,shuly:phd}.

From the point of view of grammar engineering, the abstract machine
approach has an additional advantage. \amalia's compiler incorporates
an algorithm, based on~\cite{samuelsson95}, for inverting grammars
(designed for parsing) into a form more suitable for generation. The
compiler then produces code for the inverted grammar, using exactly
the same machine language. Thus, the same grammar can be compiled to
two different object programs for the two different tasks. The
interpreter executes both kinds of programs in the same way -- only
the initialization of the machine's state and the format of the final
results differ. We thus obtain a uniform platform for developing
grammars serving both for parsing and for generation.

We discuss the use of abstract machine techniques for compilation in
the next section, and sketch the algorithm that inverts a grammar for
generation in Section~\ref {sec:generation}.
Section~\ref{sec:unified} explains the dual operation of the abstract
machine, and Section~\ref{sec:impl} lists some implementation details.

\section{Why abstract machines?}
\label{sec:abstract-machines} 
High-level programming languages with dynamic structures have always
been hard to develop compilers for. A common technique for overcoming
the problems involves the notion of an {\em abstract machine}. It is a
machine that, on one hand, captures the essentials of the high-level
language in its architecture and instruction set, such that compiling
from the source language to the (abstract) machine language becomes
relatively simple. On the other hand, the architecture must be simple
enough for the abstract machine language to be easily interpretable on
common machines. This technique also facilitates portable front ends
for compilers: as the machine language is abstract, it can be easily
interpreted on different (concrete) machines/platforms.

Abstract machines were used for various procedural and functional
languages, but they became prominent for logical programming languages
since the introduction of the Warren Abstract Machine
(WAM)~\cite{waren83,wam} for Prolog. While Prolog has gained a
recognition as a practical implementation of the idea of programming
in logic, a method for interpreting the declarative logical statements
was needed for such an implementation to be well-founded.  Even though
there were prior attempts to construct both interpreters and compilers
for Prolog, it was the WAM that gave the language not only a good,
efficient compiler, but, perhaps more importantly, an elegant
operational semantics.

The WAM immediately became the starting point for many compiler
designs for Prolog. The techniques it delineates serve not only for
Prolog proper, but also for constructing compilers for related
languages: parallel Prolog compilers, variants of Prolog that use
different resolution methods, extend Prolog with types or with record
structures, etc. An additional advantage of abstract machines is that
they are a useful tool in formally verifying the correctness of
compilers.

\section{Inverting grammars for generation}
\label{sec:generation} 
One of the attractions of declarative linguistic theories such as HPSG
is that a single grammar, formulated in the theory, can be used both
for parsing and for generation. While this is true in theory, not many
practical implementations of linguistic formalisms support
bidirectional grammar processing. Many advantages of bidirectional
natural language systems are listed in~\cite{reversible-grammar},
where three options for {\em reversibility} are considered (pp.\
xiii-xxi): (1) A grammar is compiled into two separate programs,
parser and generator, requiring a different evaluation strategy; (2)
The parser and the generator are separate programs, executed using the
same evaluation strategy; (3) The parser and the generator are one
program, and the evaluation strategy can handle it being run in either
direction.  Our solution falls into the second category: there is only
one input grammar, which is compiled into two different (abstract
machine) object programs; these two programs are executed using
exactly the same mechanism, the interpreter, and hence employ the same
strategy. This guarantees both ease of grammar development and
maintenance and no loss of efficiency.

Grammars are usually oriented towards the analysis of a string and not
towards generation from a (usually nested) semantic form. In other
words, rules reflect the phrase structure and not the
predicate-argument structure. It is therefore desirable to transform
the grammar in order to enable systematic reflection of any given
logical form in the productions. To this end we apply an {\em
inversion\/} procedure, based upon\footnote{Samuelsson's inversion
algorithm was developed for definite clause grammars
\cite{pereira80}. We ported it to a typed feature-structure
framework.}~\cite{samuelsson95}, to render the rules with the nested
predicate-argument structure, corresponding to that of input logical
forms. Once the grammar is inverted, the generation process can be
directed by the input semantic form; elements of the input are
consumed during generation just like words are consumed during
parsing.

Figure~\ref{fig:orig-gra} depicts a simple example grammar in \ale\
format\footnote{The signature is omitted for lack of space.} ({\tt
prd} stands for predicate, {\tt a} for argument, {\tt var} for
variable, {\tt rst} for restriction and {\tt conn} for
connective). The first rule creates a sentence (S) out of a noun
phrase (NP) and a verb phrase (VP). The semantics of the S (denoted by
the variable {\tt R6}) is obtained by applying the semantics of the NP
($\lambda R5.R6$) to that of the VP. In the same way, the second rule,
combining a determiner (DET) with a noun (N) to obtain an NP, applies
the meaning of the DET to that of the N to obtain (after two
$\beta$-reductions that are incorporated into the rule) the meaning of
the NP. The lexical entries of three words are shown as well.
\begin{figure*}[hbt]
{\small
\begin{verbatim}
(phrase, syn:(syn, cat:s), sem:(R6, sem))
===>
cat> (phrase, syn:(syn, cat:np), sem:(lambda, (var:R5, rst:(R6, funct)))), % head
cat> (phrase, syn:(syn, cat:vp), sem:(lambda, (var:R7, rst:(R5, funct)))).

(phrase, syn:(syn, cat:np), sem:(R6, sem))
===>
cat> (phrase, syn:(syn, cat:det), sem:(lambda, (var:R5, rst:(R6, funct)))), % head
cat> (phrase, syn:(syn, cat:n), sem:(lambda, (var:R7, rst:(R5, funct)))).

every --->
(word, syn:(syn, cat:det),
 sem:(lambda, var:R5, 
      rst:(lambda, var:R6, 
           rst:(prd:(forall, var:R2, form:(bool, conn:if, 
                                                 wff1:(R5, a1:R2),
                                                 wff2:(R6, a1:R2))),
                a1:R5, a2:R6)))).
boy --->
(word, syn:(syn, cat:n), sem:(lambda, var:R5, rst:(prd:boy, a1:R5))).

sleeps --->
(word, syn:(syn, cat:vp), sem:(lambda, var:R5, rst:(prd:sleep, a1:R5))).
\end{verbatim}
}
\caption{A simple grammar}
\label{fig:orig-gra}
\end{figure*}

Figure~\ref{fig:inv-gra} depicts (part of) the same grammar after
inversion. The inverted grammar reflects the semantic argument
structure, not the phrase structure. For example, the first rule
creates a sentence, whose {\em sem\/} feature corresponds to $\forall
R5.(R8(R5) \rightarrow R10(R5))$, from three components: an N ($R8$),
a VP ($R10$) and a {\em semantic head}, $R3$. The string generated by
the S, encoded as the value of the {\em str\/} feature (see below), is
the concatenation of the strings generated by the head, the N and the
VP. For such rules to be applicable, the lexicon has to be inverted,
too: the \mbox{``words''} of the inverted grammar are atomic semantic
formulae. The last three rules add syntactic information to the
semantics encoded in the primitives. In addition to these inverted
rules, a {\em semantic knowledge base\/} is generated,
associating semantic primitives with words. It is used in the final
stage of the generation, when the actual words are generated.
\begin{figure*}[hbt]
{\small
\begin{verbatim}
(phrase, syn:(syn,cat:s), str:[R3,R19,R22],
         sem:(R3, prd:(forall, var:R5, form:(conn:if, 
                                             wff1:(R8,a1:R5), 
                                             wff2:(R10,a1:R5))),
                  a1:R8, a2:R10))
===>
(phrase, syn:cat:n, sem:(lambda, rst:R8), str:R19),
(phrase, syn:cat:vp, sem:(lambda, rst:R10), str:R22),
(lambda, var:R8, rst:(lambda, var:R10, rst:R3)).

(word, syn:cat:n, sem:R3, str:[R5])
===>
(R3, lambda, var:R4, rst:(R5, prd:noun, a1:R4)).

(word, syn:cat:vp, sem:R3, str:[R5])
===>
(R3, lambda, var:R4, rst:(R5, prd:v_intrans, a1:R4)).

(word, syn:cat:det, sem:R3, str:[R10])
===>
(R3, lambda, var:(R4,a1:R6), 
             rst:(lambda, var:(R8, a1:R6), 
                          rst:(R10,prd:(forall, var:R6, form:(conn:if, wff1:R4, wff2:R8)),
                                   a1:R4, a2:R8))).
\end{verbatim}
}
\caption{The inverted grammar (partial)}
\label{fig:inv-gra}
\end{figure*}

Grammars must satisfy certain requirements in order for them to be
invertible. However, the requirements are not overly restrictive and
allow encoding of a variety of natural language grammars. In
particular, the semantics must be encoded by predicate-argument
structures. What the inversion in fact achieves is restructuring of a
grammar; this enables effective treatment of the nested structure of
logical forms, so that the resulting grammar is inherently suitable
for generation.

Grammar inversion is performed as part of the compilation. The given
grammar is enhanced in a way that will ultimately enable to
reconstruct the words spanned by the semantic forms. To achieve this
aim, each rule constituent is extended by an additional
special-purpose feature ({\em str\/} in the example grammar). The
value of this feature for the rule's head is set to the concatenation
of its values in the body constituents, to reflect the original phrase
structure of the rule.

Among the other advantages of the abstract machine approach mentioned
above, this technique gives an express solution for the termination
problem. It is usually difficult to define when generation terminates,
but once the query is given as a sequence of semantic components, they
are consumed in a linear manner. While generation, just like parsing,
is not guaranteed to terminate, the termination criteria of parsing
apply for our generation scheme. In other words, generation in our
system can be viewed as parsing (`consuming') input sequences of
meaning components.

\section{Unified parsing and generation}
\label{sec:unified}
\amalia\ employs a bottom-up chart based control unit, where rules are
evaluated from left to right. The chart is used for storing active and
complete edges. The latter are represented as pointers to feature
structures; the former consist of a sequence of such pointers (for the
part of the edge prior to the dot) and a pointer to the compiled code
(for the part succeeding the dot). For parsing, edges span a
sub-sequence of the input string, assigning it some structure. For
generation, edges span a sub-form of the input semantic form, also
assigning it a structure that eventually determines a phrase whose
meaning is that sub-form. It must be noted that at run-time there is
no notion of the particular task (parsing/generation) performed by the
machine, and the effect of the machine instructions is the same for
both tasks.

\begin{figure}[hbt]
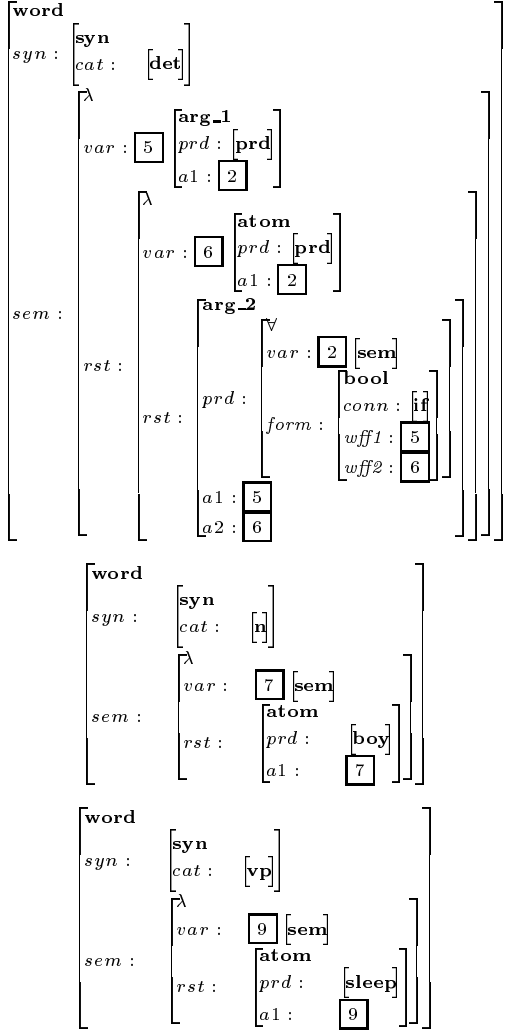

{\scriptsize
\[
\begin{array}{c}
\begin{tfs}{word}
  syn: \begin{tfs}{syn} cat: & \begin{tfs}{det} \end{tfs} \end{tfs}\\
  sem: \begin{tfs}{$\lambda$}
    var: \tag{5}\begin{tfs}{arg\_1}
              prd: \begin{tfs}{prd} \end{tfs}\\
              a1: \tag{2}
            \end{tfs}\\
    rst: \begin{tfs}{$\lambda$}
      var: \tag{6}\begin{tfs}{atom}
              prd: \begin{tfs}{prd} \end{tfs}\\
              a1: \tag{2}
            \end{tfs}\\
      rst: \begin{tfs}{arg\_2}
        prd: \begin{tfs}{$\forall$}
          var: \tag{2}\begin{tfs}{sem} \end{tfs}\\
          \mbox{\em form}: \begin{tfs}{bool}
            conn: \begin{tfs}{if} \end{tfs}\\
            \mbox{\em wff1}: \tag{5}\\
            \mbox{\em wff2}: \tag{6}
          \end{tfs} %forall
        \end{tfs} \\%bool
        a1: \tag{5}\\
        a2: \tag{6}
      \end{tfs} %atom
    \end{tfs} %lambda
  \end{tfs}\\ %lambda
\end{tfs} %word
\\ \\
\begin{tfs}{word}
  syn: & \begin{tfs}{syn} cat: & \begin{tfs}{n} \end{tfs} \end{tfs}\\
  sem: & \begin{tfs}{$\lambda$}
    var: & \tag{7}\begin{tfs}{sem} \end{tfs}\\
    rst: & \begin{tfs}{atom}
      prd: & \begin{tfs}{boy} \end{tfs}\\
      a1: & \tag{7}
    \end{tfs} %atom
  \end{tfs}\\ %\lambda
\end{tfs} %word
\\ \\
\begin{tfs}{word}
  syn: & \begin{tfs}{syn} cat: & \begin{tfs}{vp} \end{tfs} \end{tfs}\\
  sem: & \begin{tfs}{$\lambda$}
    var: & \tag{9}\begin{tfs}{sem} \end{tfs}\\
    rst: & \begin{tfs}{atom}
      prd: & \begin{tfs}{sleep} \end{tfs}\\
      a1: & \tag{9}
    \end{tfs} %atom
  \end{tfs}\\ %\lambda
\end{tfs} %word
\end{array}
\]
}
\caption{Initial chart entries, parsing}
\label{fig:init-f}
\end{figure}

\amalia's operation for generation differs from parsing only in
initialization and interpretation of the results.  For parsing, the
input is a string of words. Each word is looked up in the lexicon, and
its associated feature structure (or feature structures, in case the
word is ambiguous) is entered in the main diagonal of the chart as a
complete edge.  Thus, for the example grammar and the input ``every
boy sleeps'', the items in the $[0,1],[1,2],[2,3]$ entries of the
chart are as depicted in~Figure~\ref{fig:init-f}.

\begin{figure}[hbt]
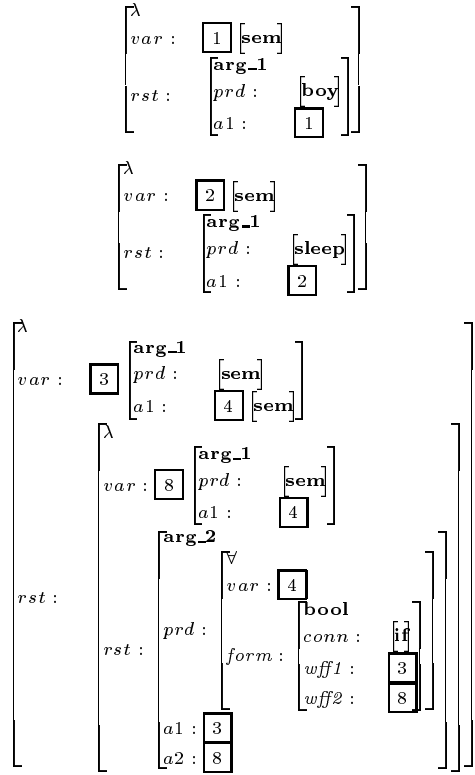

{\scriptsize
\[
\begin{array}{c}
\begin{tfs}{$\lambda$}
  var: & \tag{1}\begin{tfs}{sem} \end{tfs}\\
  rst: & \begin{tfs}{arg\_1}
    prd: & \begin{tfs}{boy} \end{tfs}\\
    a1: & \tag{1}
  \end{tfs}
\end{tfs}
\\ \\
\begin{tfs}{$\lambda$}
  var: & \tag{2}\begin{tfs}{sem} \end{tfs}\\
  rst: & \begin{tfs}{arg\_1}
    prd: & \begin{tfs}{sleep} \end{tfs}\\
    a1: & \tag{2}
  \end{tfs}
\end{tfs}
\\ \\
\begin{tfs}{$\lambda$}
  var: & \tag{3}\begin{tfs}{arg\_1}
    prd: & \begin{tfs}{sem} \end{tfs}\\
    a1: & \tag{4}\begin{tfs}{sem} \end{tfs}
  \end{tfs}\\ %arg1
  rst: & \begin{tfs}{$\lambda$}
    var:  \tag{8}\begin{tfs}{arg\_1}
      prd: & \begin{tfs}{sem} \end{tfs}\\
      a1: & \tag{4}
    \end{tfs}\\ %arg1
    rst:  \begin{tfs}{arg\_2}
      prd:  \begin{tfs}{$\forall$}
        var:  \tag{4}\\
        \mbox{\em form}:  \begin{tfs}{bool}
          conn: & \begin{tfs}{if} \end{tfs}\\
          \mbox{\em wff1}: & \tag{3}\\
          \mbox{\em wff2}: & \tag{8}
        \end{tfs} %bool
      \end{tfs}\\ % forall
      a1:  \tag{3}\\
      a2:  \tag{8}
    \end{tfs} %arg_2
  \end{tfs} % rst
\end{tfs} %lambda
\end{array}
\]
}
\caption{Initial chart entries, generation}
\label{fig:init-g}
\end{figure}

For generation, the input is a semantic form, represented as (an \ale\
description of) a feature structure. The chart is initialized with
(complete) edges that correspond to elements in the input semantic
form, rather than to words. For example, if the input is (a feature
structure encoding of) $\forall x(\mbox{\em boy}(x) \rightarrow
\mbox{\em sleep}(x))$, the items in the $[0,1],[1,2],[2,3]$ entries of
the chart are as depicted in~Figure~\ref{fig:init-g}. The first item
encodes $\lambda x.boy(x)$; the second -- $\lambda x.sleep(x)$; and
the third -- $\lambda P. \lambda Q. \forall x (P(x) \rightarrow
Q(x))$.

It must be clear that there doesn't have to be a $1-1$ correspondence
between the initial states of the chart in both tasks. The semantic
input is scanned and its elements are (recursively) selected in a
pre-defined order that is induced by the restructuring of the grammar
rules (in particular, arguments precede the predicate).

Once the chart is initialized, the same processing strategy is applied
independently of the task: the compiled program is executed on the
input. The basic operation performed by the object programs is
unification, which is needed for both tasks. Unification implements
the {\em dot movement\/} operation that lies in the heart of
chart-based parsing and generation. However, dot movement is
interpreted differently for both tasks, since the (compiled) grammar
rules are different: for parsing, dot movement goes over a sub-part of
the input phrase; for generation, it covers a part of the input
logical form.

\begin{figure}[hbt]
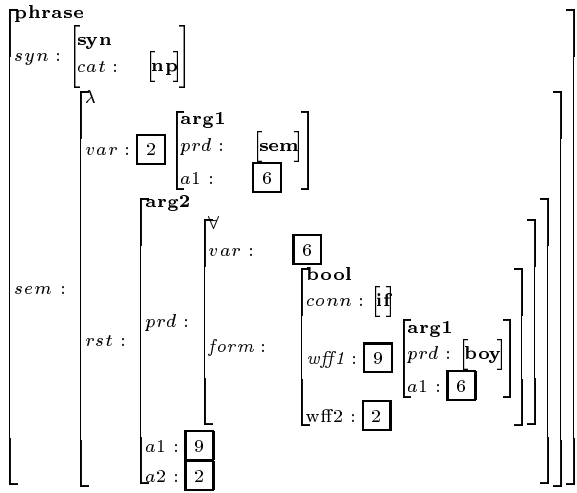

{\scriptsize
\[
\begin{tfs}{phrase}
  syn: \begin{tfs}{syn} cat: & \begin{tfs}{np} \end{tfs} \end{tfs}\\
  sem: \begin{tfs}{$\lambda$}
    var: \tag{2}\begin{tfs}{arg1}
      prd: & \begin{tfs}{sem} \end{tfs}\\
      a1: & \tag{6}
    \end{tfs}\\ %lambda
    rst: \begin{tfs}{arg2}
      prd: \begin{tfs}{$\forall$}
        var: & \tag{6}\\
        \mbox{\em form}: & \begin{tfs}{bool}
          conn: \begin{tfs}{if} \end{tfs}\\
          \mbox{\em wff1}: \tag{9}\begin{tfs}{arg1}
            prd: \begin{tfs}{boy} \end{tfs}\\
            a1: \tag{6}
          \end{tfs}\\ %arg1
          \mbox{wff2}: \tag{2}
        \end{tfs}\\ %bool
      \end{tfs}\\ %forall
      a1: \tag{9}\\
      a2: \tag{2}
    \end{tfs} %arg2
  \end{tfs} %lambda
\end{tfs}% phrase          
\]
}
\caption{Parsing (intermediate) result}
\label{fig:res-parse}
\end{figure}

Consider the effect of dot movement for parsing: assume that an active
edge corresponding to the second rule with the dot in the initial
position is applied to the lexical entry of ``every'', present in
$[0,1]$. The compiled code of the second rule is executed on
``every''; some trivial unifications take place, but the more
interesting ones bind {\tt R5} of the rule to the value of the tag
$\tag{5}$ in the lexical entry, and {\tt R6} -- to the value of the
path {\em sem:rst}. A new active edge is created, with these bindings
recorded, and entered in $[0,1]$.  The part of the edge following the
dot points to the second category in the body of this rule. Assume
further that this edge is combined with (the complete edge that is)
the lexical entry of ``boy''. Several trivial unifications take place,
but the interesting ones bind {\tt R7} in the rule to the tag \tag{7}
in ``boy'', and {\tt R5} of the rule to the value of {\em sem:rst\/}
in ``boy''.  Due to reentrancies among the rule's constituents, the
obtained (complete) edge (spanning $[0,2]$), whose {\em sem\/} feature
indeed encodes the semantics of ``every boy'' ($\lambda Q.\forall
x(boy(x) \rightarrow Q(x))$), is as depicted in
Figure~\ref{fig:res-parse}.

\begin{figure}[hbt]
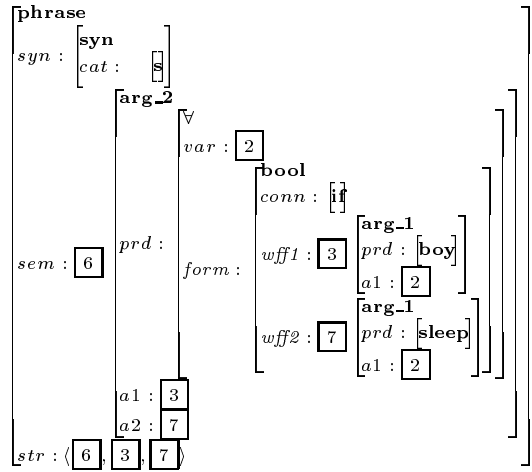

{\scriptsize
\[
\begin{tfs}{phrase}   
  syn: \begin{tfs}{syn} cat: & \begin{tfs}{s} \end{tfs} \end{tfs}\\
  sem: \tag{6}\begin{tfs}{arg\_2}
    prd: \begin{tfs}{$\forall$}
      var: \tag{2}\\
      \mbox{\em form}: \begin{tfs}{bool}
        conn: \begin{tfs}{if} \end{tfs}\\
        \mbox{\em wff1}: \tag{3}\begin{tfs}{arg\_1}
          prd: \begin{tfs}{boy} \end{tfs}\\
          a1: \tag{2}
        \end{tfs}\\ %arg_1
        \mbox{\em wff2}: \tag{7}\begin{tfs}{arg\_1}
          prd: \begin{tfs}{sleep} \end{tfs}\\
          a1: \tag{2}
        \end{tfs} %arg_1
      \end{tfs} %bool
    \end{tfs}\\ %forall
    a1: \tag{3}\\
    a2: \tag{7}
  \end{tfs}\\ %arg_2
  str: \langle \tag{6},\tag{3},\tag{7} \rangle
\end{tfs}% phrase
\]
}
\caption{Generation result}
\label{fig:res-gen}
\end{figure}

Next, we give a scenario of a generation process. It is easy to see
how the last three rules of the inverted grammar are applicable to the
three lexical entries of Figure~\ref{fig:init-g}, respectively. Assume
an active edge corresponding to the first rule is present in $[0,0]$,
with the dot in the initial position. Two dot movements, over the
first two elements in the body of this rule, bind {\tt R8} to the
value of {\em rst\/} in the lexical entry of $\lambda(x).boy(x)$, and
{\tt R10} -- to the value of {\em rst\/} in $\lambda(x).sleep(x)$. An
active edge, with the dot in the penultimate position, is obtained in
$[0,2]$. The next dot movement applies (the code that was generated
for) the last body element of the rule to the lexical entry residing
in $[2,3]$. {\tt R8} of the rule is unified with the value of the tag
\tag{3} in this entry; since {\tt R8} was bound by previous
unifications, the value of {\em prd\/} is set to {\em boy}. {\tt R10}
of the rule is unified with the value of \tag{8}, and the second
predicate is set to {\em sleep}. Finally, {\tt R3} is unified with the
value of {\em rst:rst\/} in the lexical entry; the complete edge
created, spanning the entire input, is depicted in
Figure~\ref{fig:res-gen}.

The chart algorithm ends up with a (possibly empty) set of feature
structures, spanning the entire input: these are all the complete
edges derivable from the input and the grammar rules (there is no
notion of an {\em initial symbol\/}). Of course, if the grammar is
such that an infinite number of derivations can be produced,
computations might not terminate (\amalia\ does not incorporate a
subsumption check to test for spurious ambiguity).  For parsing, the
results depict different structures of the input string. Ideally, they
contain some representation of the string's semantics. This is also
true for generation, with a slight difference: according to the
grammar inversion algorithm, each resultant structure is guaranteed to
have a feature (namely, {\em str}) that encodes a list of words,
comprising the phrase generated. As can be seen in the example
(Figure~\ref{fig:res-gen}), the value of this feature is not a list of
words but rather a list of feature structures, each of which
corresponds to (i.e., is subsumed by) a lexical entry in the inverted
grammar. A final post-processing stage generates all the possible
strings using this list and the semantic knowledge base.

\section{Implementation}
\label{sec:impl}
This section describes the input language for \amalia\ grammars and
touches on some implementation details. In particular, it discusses
the differences between \amalia\ and \ale\ in terms of expressiveness
and efficiency.

\amalia\ supports the same type hierarchies as \ale\ does, with
exactly the same specification syntax. This means that the user can
specify any bounded-complete partial order as the type hierarchy. Only
immediate sub-types are specified, and the reflexive-transitive
closure of the sub-type relation is computed automatically by the
compiler. The special type {\tt bot} must be declared as the unique
most general type.

Appropriateness, too, is specified using \ale's syntax, by listing
features at the type they are introduced by. The feature introduction
condition must be obeyed: every feature must be introduced by some
most general type, and is appropriate for all its sub-types. However,
\amalia\ allows appropriateness loops\footnote{Appropriateness loops
are handled by employing lazy evaluation techniques at run-time.} in
the type hierarchy. Type constraints are not supported by \amalia.

\amalia\ uses a subset of \ale's syntax for describing feature
structures. As a rule, whenever \amalia\ supports \ale's
functionality, it uses the same syntax. In general, \amalia\ supports
totally well-typed, possibly cyclic, non-disjunctive feature
structures. Set values, as in \ale, are not supported, but list values
are. \amalia\ does not respect the distinction between {\em
intensional\/} and {\em extensional\/} types~\cite[Chapter
8]{carp92}.  Also, feature structures cannot incorporate inequality
constraints.

The semantics of the logical descriptions, as well as the operator
precedence, follow \ale. As in \ale, partial descriptions are expanded
at compilation time. \amalia's compiler performs type inference on
partial descriptions, reports any inconsistencies, and then creates
code for the expanded structures. To avoid infinite processing in the
face of appropriateness loops (where no finite totally well-typed
structure that satisfies the description might exist), the compiler
stops expanding a structure if it is the most general structure of its
type.

\ale\ includes a built-in definite logic programming language;
\amalia\ does not. The entire power of definite clause specifications
is missing in \amalia. However, a few common functions that are
external to the feature structure formalism were added to the system,
and grammar specifications can use them. These features are referred
to as {\em goals\/}, although it must be remembered that they are far
weaker than \ale's goals.

\amalia\ preserves \ale's syntax in describing lexical
entries. Multiple lexical entries may be provided for each word,
separated by semicolons. It also keeps \ale's syntax in the definition
of {\em empty categories\/} (or $\epsilon$-rules). In contrast to
\ale, \amalia\ processes empty categories at compile time. Each empty
category is matched by the compiler against each element in the body
of every rule; if the unification succeeds, a new rule is added to the
grammar, based upon the original rule, with the matched element
removed. Some limitations apply for this process (which in the general
case is not guaranteed to terminate), and therefore the resulting
grammar might not be equivalent to the original one.

\amalia\ supports macros in a similar way to \ale. The syntax is the
same, and macros can have parameters or call other macros (though not
recursively, of course). \ale's special macros for lists are supported
by \amalia.  Lexical rules are not supported in this version of
\amalia. \amalia's syntax for phrase structure rules is similar to
\ale's, with the exception of the \verb|cats>| specification
(permitting a list of categories in the body of a rule) which is not
supported.

The design details of the abstract machine are outside the scope of
this paper; the reader is referred to~\cite{shuly:nlulp-95,shuly:phd}
for more information on the machine itself and to~\cite{gabr:thesis}
for a detailed description of the grammar inversion. A practical
description of \amalia, its deviations from \ale\ and a complete
user's guide, are given in~\cite{amalia-man}. 

\amalia\ is implemented in $C$, augmented by {\em yacc}, {\em lex} and
{\em Tcl/Tk}~\cite{tcltk}. It was tested on various Unix platforms and
on IBM PCs.  Two versions of \amalia\ exist: an interactive,
easy-to-use, graphically interfaced system and a text-oriented,
non-interactive one. The former is intended for developing prototype
grammars; the latter is far more efficient but less user-friendly, and
is intended to be used for batch processing.  In addition, the system
functions as a graphical development framework for grammar engineers
by providing some tracing and debugging options.  The user can direct
the system to execute a program in its entirety, to break at a certain
instruction or to proceed in steps, stopping after each executed
instruction. Throughout the process of grammar execution, the abstract
machine's internal state is displayed for the user to inspect. The
main data structure upon which feature structures are being built, the
{\em heap}, is displayed, along with the machine's general-purpose and
special-purpose {\em registers}. Moreover, the contents of the chart
can be graphically displayed at any time and the derived structure can
be recovered. Grammar development becomes an easier, simpler process.

The system was tested with a wide variety of grammars, mostly
adaptations of existing \ale\ grammars.  While most of the example
grammars are rather small, we believe that the system can handle
real-scale grammar quite efficiently; however, to accommodate large
type hierarchies some major space optimizations must be introduced.
It is important to emphasize that \amalia\ does not provide the wealth
of input specifications \ale\ does. On the other hand, development of
grammars in \amalia\ is made easier due to the GUI and the improved
performance over \ale. The support of generation is unique to our
system.

To compare \amalia\ with \ale\ we have used a few benchmark
grammars. The first is an early version of an HPSG-based Hebrew
grammar described in~\cite{shuly:phd}. It consists of 4 rules and one
empty category; the type hierarchy contains 84 types and 32 features,
and the lexicon contains 13 words. The second is an HPSG-based grammar
for a subset (emphasizing relative clauses) of the Russian language
described in \cite{russian-grammar}. It consists of 8 rules and
76 lexical entries; the type hierarchy contains 151 types and 31
features. The third example is a simple grammar generating the
language $\{a^nb^n\mid n>0\}$.  Both systems were used to compile the
same grammar and to parse the same strings. The results of a
performance comparison of \amalia\ and \ale\ are listed in Figure~\ref
{fig:performance} (all times are in seconds; $n$ indicates the input
string's length and $r$ -- the number of results).  While the
execution times for the last grammar are less impressing, the
differences in compilation time indicate a major advantage in using
\amalia\ for instructional purposes; in such cases grammars are
compiled over and over again, while they are usually executed only a
few times. Limited experiments we have conducted reveal that
generation (as well as compilation for generation) is 40\%--100\%
slower than parsing (we do not know of good benchmarks for generation).

\begin{figure}[hbt]
\center
{\footnotesize
\begin{tabular}{|l||r|r|} \hline
task    & \multicolumn{1}{l|}{\ale} & \multicolumn{1}{l|}{\amalia} \\ \hline
\multicolumn{3}{|c|}{Grammar 1} \\ \hline
Compilation                     & 35.0  & 1.4   \\ \hline
Parsing, n=6, r=2     & 0.5   & 0.5   \\ \hline
Parsing, n=10, r=8    & 3.2   & 0.8   \\ \hline
Parsing, n=14, r=125  & 140.0 & 9.0   \\ \hline
\multicolumn{3}{|c|}{Grammar 2} \\ \hline
Compilation                     & 68.0  & 2.3   \\ \hline
Parsing, n=2, r=2     & 0.5   & 0.8   \\ \hline
Parsing, n=4, r=2     & 2.4   & 0.9   \\ \hline
Parsing, n=7, r=2     & 5.1   & 1.1   \\ \hline
Parsing, n=8, r=2     & 7.8   & 1.2   \\ \hline
Parsing, n=12, r=2    & 17.0  & 1.5   \\ \hline
\multicolumn{3}{|c|}{Grammar 3} \\ \hline
Compilation     & 6.5   & 0.2   \\ \hline
Parsing, n=4    & 0.1   & 0.2   \\ \hline
Parsing, n=8    & 0.8   & 0.3   \\ \hline
Parsing, n=16   & 2.8   & 1.1   \\ \hline
Parsing, n=32   & 26.0  & 16.0  \\ \hline
\end{tabular}
}
\caption{Performance comparison of \ale\ and \amalia}
\label{fig:performance}
\end{figure}

\begin{scriptsize}
% \bibliography{nlp}

\begin{thebibliography}{A\"{\i}t-Kaci \& Podelski 93}

\bibitem[A\"{\i}t-Kaci \& Podelski 93]{life-meaning}
(A\"{\i}t-Kaci \& Podelski 93)
Hassan A\"{\i}t-Kaci and Andreas Podelski.
\newblock Towards a meaning of {LIFE}.
\newblock {\em Journal of Logic Programming}, 16(3-4):195--234, July-August
  1993.

\bibitem[A\"{\i}t-Kaci 91]{wam}
(A\"{\i}t-Kaci 91)
Hassan A\"{\i}t-Kaci.
\newblock {\em Warren's Abstract Machine: A Tutorial Reconstruction}.
\newblock Logic Programming. The {MIT} Press, Cambridge, Massachusetts, 1991.

\bibitem[Carpenter \& Penn 95]{carp95}
(Carpenter \& Penn 95)
Bob Carpenter and Gerald Penn.
\newblock Compiling typed attribute-value logic grammars.
\newblock In Harry Bunt and Masaru Tomita, editors, {\em Current Issues in
  Parsing Technologies}, volume~2. Kluwer, 1995.

\bibitem[Carpenter 92a]{ale}
(Carpenter 92a)
Bob Carpenter.
\newblock {ALE} -- the attribute logic engine: User's guide.
\newblock Technical report, Laboratory for Computational Linguistics,
  Philosophy Department, Carnegie Mellon University, Pittsburgh, PA 15213,
  December 1992.

\bibitem[Carpenter 92b]{carp92}
(Carpenter 92b)
Bob Carpenter.
\newblock {\em The Logic of Typed Feature Structures}.
\newblock Cambridge Tracts in Theoretical Computer Science. Cambridge
  University Press, 1992.

\bibitem[Erbach 94]{profit}
(Erbach 94)
Gregor Erbach.
\newblock {ProFIT}: Prolog with features, inheritance and templates.
\newblock {CLAUS} Report~42, Computerlinguistik, Universit\"at des Saarlandes,
  D-66041, Saarbr\"ucken, Germany, July 1994.

\bibitem[Franz 90]{franz}
(Franz 90)
Alex Franz.
\newblock A parser for {HPSG}.
\newblock Report {CMU-LCL-90-3}, Laboratory for Computational Linguistics,
  Department of Philosophy, Carnegie Mellon University, Pittsburgh, PA 15213,
  July 1990.

\bibitem[Gabrilovich \& Estrin 96]{russian-grammar}
(Gabrilovich \& Estrin 96)
Evgeniy Gabrilovich and Arkady Estrin.
\newblock An {HPSG} grammar for the {R}ussian language.
\newblock To appear as a technical report, Laboratory for Computational
  Linguistics, the Technion, 1996.

\bibitem[Gabrilovich 97]{gabr:thesis}
(Gabrilovich 97)
Evgeniy Gabrilovich.
\newblock Natural language generation by abstract machine.
\newblock M.Sc. thesis, Technion, Israel Institute of Technology,
  Haifa, Israel, 1997.
\newblock In preparation.

\bibitem[Haddock {\it et al.} 87]{cg}
(Haddock {\it et al.} 87)
Nicholas Haddock, Ewan Klein, and Glyn Morill, editors.
\newblock {\em Categorial Grammar, Unification and Parsing}, volume~1 of {\em
  Working Papers in Cognitive Science}.
\newblock University of Edinburgh, Center for Cognitive Science, 1987.

\bibitem[Kaplan \& Bresnan 82]{lfg}
(Kaplan \& Bresnan 82)
R.~Kaplan and J.~Bresnan.
\newblock Lexical functional grammar: A formal system for grammatical
  representation.
\newblock In J.~Bresnan, editor, {\em The Mental Representation of Grammatical
  Relations}, pages 173--281. MIT Press, Cambridge, Mass., 1982.

\bibitem[Ousterhout 94]{tcltk}
(Ousterhout 94)
John~K. Ousterhout.
\newblock {\em {Tcl} and the {Tk} Toolkit}.
\newblock Addison-Wesley Professional Computing Series. Addison-Wesley, 1994.

\bibitem[Pereira \& Warren 80]{pereira80}
(Pereira \& Warren 80)
Fernando C.~N. Pereira and David H.~D. Warren.
\newblock Definite clause grammars for language analysis -- a survey of the
  formalism and a comparison with augmented transition networks.
\newblock {\em Artificial Intelligence}, 13:231--278, 1980.

\bibitem[Pollard \& Sag 94]{hpsg2}
(Pollard \& Sag 94)
Carl Pollard and Ivan~A. Sag.
\newblock {\em Head-Driven Phrase Structure Grammar}.
\newblock University of Chicago Press and CSLI Publications, 1994.

\bibitem[Prudian \& Pollard 85]{prupol85}
(Prudian \& Pollard 85)
Derek Prudian and Carl Pollard.
\newblock Parsing head-driven phrase structure grammar.
\newblock In {\em Proceedings of the 23rd Annual Meeting of the Association for
  Computational Linguistics}, Chicago, IL., 1985. University of Chicago.

\bibitem[Samuelsson 95]{samuelsson95}
(Samuelsson 95)
Christer Samuelsson.
\newblock An efficient algorithm for surface generation.
\newblock In {\em Proceedings of the International Joint Conference on
  Artificial Intelligence}, 1995.

\bibitem[Strzalkowski 94]{reversible-grammar}
(Strzalkowski 94)
Tomek Strzalkowski, editor.
\newblock {\em Reversible Grammar in Natural Language Processing}.
\newblock The {K}luwer International Series in Engineering and Computer
  Science. Kluwer Academic Publishers, 1994.

\bibitem[Warren 83]{waren83}
(Warren 83)
David H.~D. Warren.
\newblock An abstract {P}rolog instruction set.
\newblock Technical Note 309, {SRI} International, Menlo Park, CA., August
  1983.

\bibitem[Wintner \& Francez 95]{shuly:nlulp-95}
(Wintner \& Francez 95)
Shuly Wintner and Nissim Francez.
\newblock An abstract machine for typed feature structures.
\newblock In {\em Proceedings of the 5th Workshop on Natural Language
  Understanding and Logic Programming}, pages 205--220, Lisbon, May 1995.

\bibitem[Wintner 97]{shuly:phd}
(Wintner 97)
Shuly Wintner.
\newblock {\em An Abstract Machine for Unification Grammars}.
\newblock PhD thesis, Technion -- Israel Institute of Technology,
  Haifa, Israel, January 1997.

\bibitem[Wintner {\it et al.} 97]{amalia-man}
(Wintner {\it et al.} 97)
Shuly Wintner, Evgeniy Gabrilovich, and Nissim Francez.
\newblock {AMALIA} -- {A}bstract {MA}chine for {LI}inguistic {A}pplications --
  user's guide.
\newblock Laboratory for Computational Linguistics, Computer Science
  Deparmtent, Technion, Israel Institute of Technology, 32000 Haifa, Israel,
  January 1997.

\end{thebibliography}

\end{scriptsize}

\end{document}